\documentclass[conference]{IEEEtran}
\IEEEoverridecommandlockouts

\usepackage{cite}
\usepackage{amsmath,amssymb,amsfonts}
\usepackage{algorithm}
\usepackage[noend]{algpseudocode}
\usepackage{graphicx}
\usepackage{textcomp}
\usepackage{xcolor}
\def\BibTeX{{\rm B\kern-.05em{\sc i\kern-.025em b}\kern-.08em
    T\kern-.1667em\lower.7ex\hbox{E}\kern-.125emX}}

\usepackage{booktabs} 
\usepackage{comment}

\usepackage{lastpage}
\usepackage{tikz}

\usepackage{array}
\usepackage{tabu}
\usepackage{hhline}

\usepackage{multirow}

\usepackage{breqn}

\usepackage{subfigure}

\usepackage{listings}
\usepackage{multirow}

\begin{document}

\title{Dynamic User-controllable Privacy-preserving Few-shot Sensing Framework}

\author{\IEEEauthorblockN{Ajesh Koyatan Chathoth}
\IEEEauthorblockA{
\textit{University of Pittsburgh}\\
Pittsburgh, PA, USA}

\and
 \IEEEauthorblockN{Shuhao Yu}
\IEEEauthorblockA{\textit{University of Pittsburgh}\\
 Pittsburgh, PA, USA} 

\and
\IEEEauthorblockN{Stephen Lee}
\IEEEauthorblockA{
\textit{University of Pittsburgh}\\
Pittsburgh, PA, USA}

}

\maketitle

User-controllable privacy is important in modern sensing systems, as privacy preferences can vary significantly from person to person and may evolve over time. This is especially relevant in devices equipped with Inertial Measurement Unit (IMU) sensors, such as smartphones and wearables, which continuously collect rich time-series data that can inadvertently expose sensitive user behaviors. While prior work has proposed privacy-preserving methods for sensor data, most rely on static, predefined privacy labels or require large quantities of private training data, limiting their adaptability and user agency. In this work, we introduce PrivCLIP, a dynamic, user-controllable, few-shot privacy-preserving sensing framework. PrivCLIP allows users to specify and modify their privacy preferences by categorizing activities as sensitive (black-listed), non-sensitive (white-listed), or neutral (gray-listed). Leveraging a multimodal contrastive learning approach, PrivCLIP aligns  IMU sensor data with natural language activity descriptions in a shared embedding space, enabling few-shot detection of sensitive activities. When a privacy-sensitive activity is identified, the system uses a language-guided activity sanitizer and a motion generation module (IMU-GPT) to transform the original data into a privacy-compliant version that semantically resembles a non-sensitive activity.
We evaluate PrivCLIP on multiple human activity recognition datasets and demonstrate that it significantly outperforms baseline methods in terms of both privacy protection and data utility.



\maketitle
\section{Introduction}

A growing number of smart devices, including wearables and smartphones, are equipped with sensors that enable applications in health monitoring, fitness tracking, and human activity recognition (HAR). Among these, inertial measurement units (IMUs) are particularly useful, as they capture fine-grained motion data that can be used to infer user behavior, physical condition, and mobility patterns. Typically, this sensor data is collected and transmitted to third-party cloud services for large-scale sensing and analytics. In many applications, online data transmission is desirable. Online tracking facilitates data sharing with peers, which enhances user engagement by providing timely feedback and positive reinforcement, which can be critical for  sustained participation.

However, outsourcing data processing to third-party providers raises significant privacy concerns. This is because IMU data may contain highly sensitive information about individuals. For example, raw accelerometer readings can inadvertently reveal sensitive user information, including physical activity patterns and health conditions~\cite{malekzadeh2018protecting}. While cloud services may not be malicious, they are often considered semi-honest: they follow the intended agreements but may still attempt to infer sensitive information from the data they process. 
Moreover, users are increasingly concerned about the potential misuse of their personal data, including the possibility that it might be sold to third parties without their consent. 
Consequently, there is a growing interest in privacy-preserving and trustworthy analysis of sensor data,  techniques that aim to derive meaningful insights while minimizing exposure of the raw sensing information~\cite{diraco2023human}.
Existing studies have primarily focused on protecting privacy by transforming user data to obscure sensitive information~\cite{raval2019olympus}. 
Leveraging deep learning techniques, these approaches learn data perturbations or add noise that reduce the risk of leaking private attributes, while still preserving utility for the target application~\cite{jain2021differentially, chathoth2021federated, chathoth2022differentially}.
Despite promising results, prior techniques impose significant limitations on user agency. In particular, they often fail to support dynamic and personalized privacy preferences. Most existing models assume static user-defined policies and train once on a fixed set of privacy labels or objectives~\cite{malekzadeh2018replacement}. This rigidity severely limits their flexibility. If a user's privacy preferences change, such as wanting to obscure a new type of sensitive activity, the entire model typically needs to be retrained or fine-tuned with updated labels, which is computationally expensive and impractical in real-world deployments.  For instance, in replacement-based privacy preservation techniques, non-sensitive activities replace sensitive ones in the feature space using an autoencoder-based technique~\cite{malekzadeh2018replacement}. While effective in specific scenarios, such methods cannot easily adapt to new privacy requirements without retraining the underlying models.

Another major challenge lies in the limited availability of data for training privacy-preserving models. In many scenarios, collecting high-quality annotated data that distinguishes between private and non-private activities is both impractical and costly. Data is often collected from a small number of users, making it difficult to capture different types of activities. This data scarcity introduces several challenges for learning privacy-preserving HAR models. Deep learning approaches generally require large amounts of labeled data to achieve robust performance, especially when simultaneously optimizing for both utility (e.g., activity recognition) and privacy (e.g., obfuscating sensitive patterns). While numerous studies have focused on learning from limited data in the general HAR domain, little work has addressed this problem concerning privacy protection~\cite{malekzadeh2018replacement}.
Designing such techniques is highly relevant, as it can substantially reduce the effort required to develop privacy-preserving models in data-constrained environments.

To address the above challenges, we propose a few-shot sensing framework, PrivCLIP, which employs contrastive learning to learn an expressive joint IMU–text representation from limited data, with the goal of protecting sensitive information embedded in raw IMU signals. By leveraging multimodal contrastive learning techniques, our framework enables the recognition of diverse human activities while maintaining privacy in low data settings. Similar to prior work~\cite{malekzadeh2018replacement}, we categorize sensor data into three groups: (i) black-listed activities that are deemed sensitive by users (ii) gray-listed activities that are neither clearly sensitive nor essential for utility, and (iii) white-listed activities necessary to support utility by applications. This categorization enables users to specify their privacy policies for IMU data, and dynamically control which types of sensor data are protected, thereby preventing third-party services from accurately inferring sensitive activities from the shared data. Our key contributions are:

\begin{itemize}
    \item We propose Priv-CLIP, a novel few-shot sensory data classification and replacement technique based on contrastive learning for time-series sensory data to preserve the privacy of user activities. Our multimodal approach augments IMU signals with textual descriptions generated from the data. This multimodal representation enables more robust activity classification, allowing sensitive activity patterns to be replaced.
    \item We introduce PrivacyPersonalizer, a system that allows users to specify a personalized list of sensitive inferences they wish to prevent. This list is used to guide the transformation of sensor data to obscure or replace the targeted inferences. Unlike prior work that relies on fixed transformations to mask predefined sensitive activities, our approach supports diverse and dynamic privacy preferences without requiring model retraining or redeployment.
    \item We evaluate our approach on multiple IMU datasets and demonstrate that it can dynamically adapt to varying privacy preferences. We show that our method can replace sensitive sensory data while maintaining the integrity of non-sensitive sensor data, thus preserving privacy without compromising utility. Compared to baseline techniques, our method consistently outperforms them across key metrics. Furthermore, we show that our approach is effective in a few-shot setting, achieving high accuracy even with as few as eight data samples.

\end{itemize}

\section{Background}
\subsection{HAR Privacy}


Inertial Measurement Units (IMUs), which typically consist of accelerometers and gyroscopes, are widely used for human activity recognition (HAR). These sensors generate multivariate time-series data that can be used to train machine learning models to classify a wide range of physical activities. However, a significant privacy concern arises from the fact that this data can also be exploited by malicious parties to infer sensitive activities that users may not wish to disclose. For example, when IMU data is transmitted to cloud-based services for processing, it leaves the user's device and becomes vulnerable to misuse. An adversary or unauthorized entity could analyze the data to infer private behaviors such as smoking or sedentary periods, even if the data was originally collected for innocuous purposes like step counting. This raises privacy concerns, as users have little to no control over which activities can be inferred from their sensor data.


A growing body of research explores the use of machine learning (ML) techniques to preserve user privacy~\cite{yang2024privacy}. In this approach, ML models are trained to transform raw sensor data in a way that filters out sensitive information, preventing its inference by third-party service providers~\cite{raval2019olympus, malekzadeh2018protecting}. However, these methods typically require access to large amounts of labeled sensitive data, which can be difficult to obtain and may raise additional privacy concerns. While prior work has explored few-shot learning approaches for activity classification in data-constrained settings~\cite{feng2019few, ganesha2024few}, there has been limited exploration of few-shot methods for privacy-preserving transformation.



\subsection{User-controllable HAR Privacy}
Users often have diverse and dynamic preferences regarding what types of information they consider sensitive. They may wish to selectively disclose data based on activity types, and these preferences can vary significantly between individuals. For example, one user may be comfortable sharing cycling activity but prefer to keep step count private, while another might choose the opposite. These preferences are also context-dependent and can shift based on the application, time, or situation—for instance, a user might allow sharing with a health app but not a social media platform, or may tighten privacy settings during travel or after experiencing a data breach. While prior studies have explored user-controllable privacy~\cite{caputo2022you}, there is limited research on how such user-controllable privacy can be realized specifically in HAR systems.

Most existing HAR privacy approaches rely on static policies, where privacy-sensitive attributes are predefined, and models are trained accordingly~\cite{malekzadeh2019mobile,malekzadeh2018replacement}. Once these models are deployed, their behavior is fixed, offering little to no flexibility to accommodate user-specific or context-dependent privacy preferences. While some recent studies have proposed conditional privacy mechanisms, these approaches still depend on a fixed set of predefined conditions, limiting their ability to adapt to the dynamic and individualized nature of user privacy expectations. 
One simple solution might be to train a personalized model for each unique privacy preference, but this approach is not scalable due to the high computational and data requirements, as well as the impracticality of anticipating every possible user scenario. Thus, there is a pressing need for adaptive, user-controllable privacy mechanisms where users can dynamically set and modify their privacy preferences, and where the underlying models can respond accordingly without requiring full retraining.

\subsection{Threat Model}

\begin{figure}[t]
    \centering
    \includegraphics[width=3.4in]{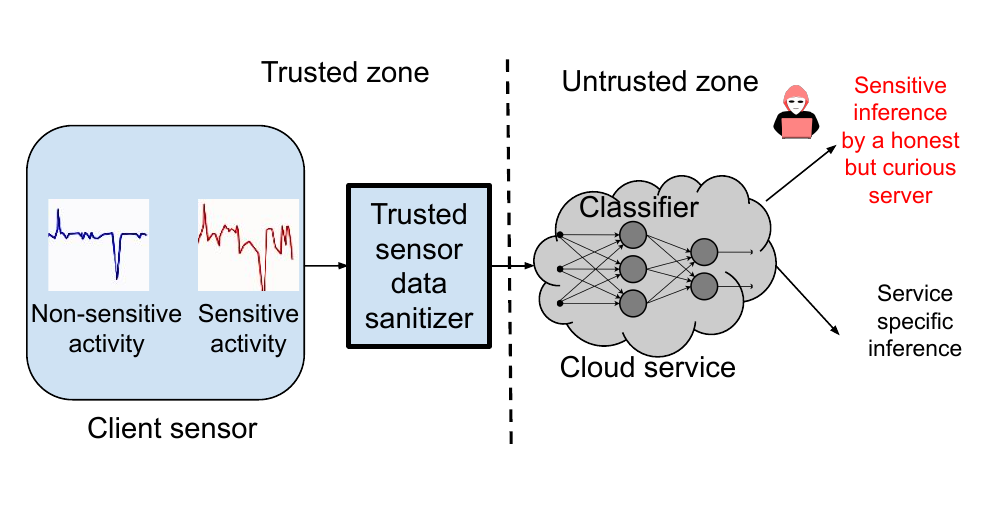}
    \caption{hreat model} 
    \label{fig:CLIP-HAR-TM}
\end{figure}

We assume an honest-but-curious threat model in which third-party cloud providers or applications are trusted to provide intended services, but may also attempt to infer sensitive information from the data they receive. This threat model is illustrated in Figure~\ref{fig:CLIP-HAR-TM}. Specifically, once sensor data is shared with a third-party service, it may be used not only for service delivery but also for unintended inferences via machine learning models.
In this setting, we also assume the presence of a trusted client-side module, such as a mobile phone, smartwatch, or edge gateway, that sits between the sensor and the third-party application. This trusted module operates in a secure environment and is responsible for masking the raw sensor data according to the user's privacy preferences before any data leaves the user's control. The goal of this trusted module is to enforce user-defined privacy controls dynamically, preventing third-party applications from learning or inferring sensitive information based on user-defined preferences. These privacy preferences are personalized and may vary from one user to another or evolve over time. Therefore, the privacy-preserving mechanism must be adaptable, allowing the trusted module to flexibly transform or replace data to meet the user’s current privacy requirements without the need to retrain or redeploy the obfuscation model.


\subsection{Problem statement}
We consider a problem setting where users aim to selectively prevent certain types of inferences from being made on their sensory data, while still enabling meaningful utility for non-sensitive information. Let $\mathcal{X}$ denote the space of raw sensory input sequences, and $\mathcal{Y}$ the set of possible inference labels (e.g., walking, sleeping). Each user specifies a personalized privacy preference in the form of a blacklist $\mathcal{Y}_{\text{private}} \subseteq \mathcal{Y}$, representing the subset of labels they wish to protect from third-party inference. We assume an honest-but-curious adversary model that correctly performs the intended services but may also attempt to infer sensitive labels in $\mathcal{Y}_{\text{private}}$ using machine learning models trained on the data it receives. The adversary only has access to transformed data $x' \in \mathcal{X}'$, where $\mathcal{X}'$ is the space of privacy-preserving representations. The objective is to design a transformation function $T: \mathcal{X} \rightarrow \mathcal{X}'$ that minimizes the adversary’s ability to infer labels in $\mathcal{Y}_{\text{private}}$ from $x'$, while still allowing accurate inference of labels in $\mathcal{Y} \setminus \mathcal{Y}_{\text{private}}$. Importantly, this transformation must dynamically respect each user’s specified privacy preferences without requiring retraining or redeployment of the model. The key challenge lies in learning such a function $T$ that achieves sensitive inference prevention and protects utility, even in few-shot learning scenarios with limited labeled data.

\section{PrivCLIP Design}

\begin{figure}[t]
    \centering
    \includegraphics[width=2.6in]{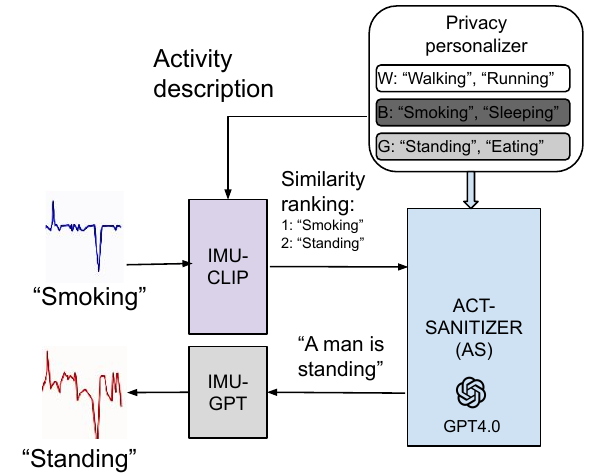} 
    \caption{PrivCLIP Architecture.
    }
    \label{fig:CLIP-HAR-Replacement}
\end{figure}
The PrivCLIP architecture, illustrated in Figure~\ref{fig:CLIP-HAR-Replacement}, consists of three key components that provide privacy-preserving activity recognition from sensory data:
(i) IMU-CLIP: This is a few-shot sensitive activity detection module built using contrastive learning. It maps raw IMU (Inertial Measurement Unit) signals into a shared embedding space where activities can be accurately recognized with minimal labeled examples. IMU-CLIP enables the identification of sensitive activities, even with limited training data, by leveraging semantic similarity between sensor sequences and activity descriptions.
(ii) Privacy Personalizer: This module provides users with fine-grained control over their privacy preferences. Users specify a personalized privacy policy in the form of a set of sensitive activity labels they wish to suppress. The Privacy Personalizer translates this user-defined list into a set of constraints that guide downstream data transformation. (iii) ACT-SANITIZER: This is the core data transformation component responsible for modifying raw IMU signals to suppress inference of the user-specified sensitive activities. It takes both IMU-CLIP's output and the user’s privacy preferences as input and outputs a transformed version of the data if it contains sensitive information. In the following sections, we detail the architecture of each component.

\subsection{IMU-CLIP} 
Autoencoder-based privacy-preserving techniques often struggle with imbalanced datasets and typically require large amounts of data to generalize effectively~\cite{10.1007/978-981-96-2468-3_16}. This limitation poses a significant challenge in privacy-preserving human activity recognition (HAR), where labeled data for sensitive activities is scarce. To address this, we introduce IMU-CLIP, a few-shot activity detection module designed to recognize activities, including sensitive activities, from IMU sensor data with minimal labeled examples. Few-shot detection is particularly well-suited for privacy-preserving settings, where collecting and labeling data for sensitive activities is not only difficult but may also raise ethical concerns~\cite{ruan2024advances}.

We implement IMU-CLIP for sensitive activity classification using a contrastive learning (CL) approach, which is particularly well-suited for scenarios with limited labeled data. Contrastive learning is a self-supervised technique that learns discriminative and generalizable representations by distinguishing between similar and dissimilar examples within a dataset. It has shown strong performance in various domains, including natural language processing and computer vision, by enabling models to learn from unlabeled data through the use of pairwise comparisons~\cite{nguyen2021contrastive, kumar2022contrastive}.

In our setting, we adapt contrastive learning to time-series data from inertial measurement units (IMUs) for human activity recognition. Specifically, we apply a contrastive loss function that pulls together the embeddings of similar activity sequences—such as different instances of walking or running—while pushing apart embeddings of dissimilar activities—such as walking and sleeping. This structured embedding space facilitates few-shot classification, allowing the model to identify sensitive activities accurately even when only a few labeled examples are available.

\begin{figure*}[t]
    \centering




    \includegraphics[width=5.3in]{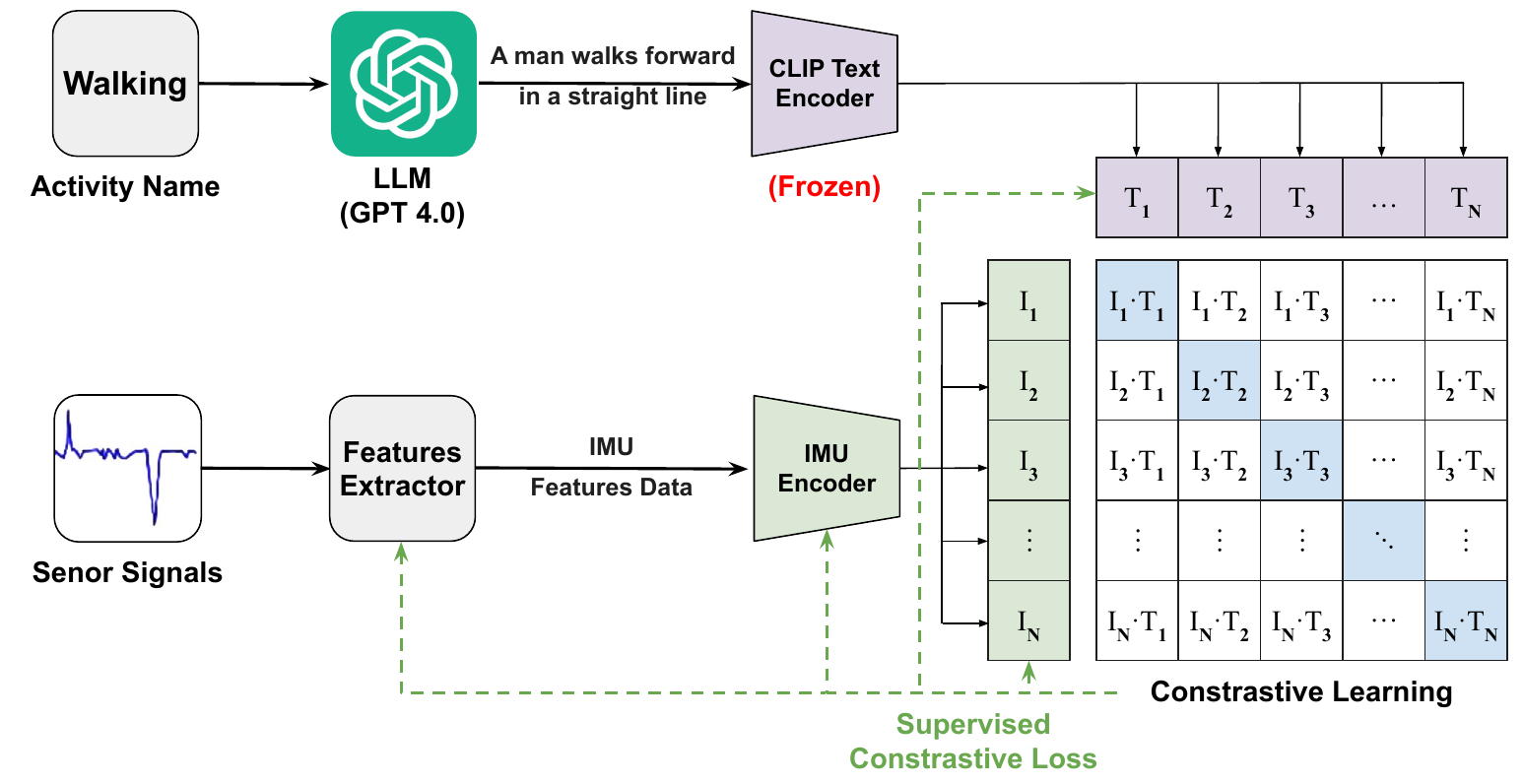} 
    \caption{IMU-CLIP architecture that learns to align IMU and text embeddings.} 
    \label{fig:CLIP-HAR}
\end{figure*}

{\bf Architecture.} We propose a multimodal contrastive learning technique, IMU-CLIP, designed to align embeddings of IMU sensory data with textual activity descriptions within a shared semantic space. This alignment leverages the similarity between learned representations of time-series sensor inputs and natural language class labels, enabling effective cross-modal understanding. As illustrated in Figure~\ref{fig:CLIP-HAR}, IMU-CLIP comprises three main components: an IMU feature extractor, an IMU encoder, and a pretrained CLIP text encoder.

\begin{figure}[t]
\small
\begin{lstlisting}[caption={Activity Description Prompt Template}, label={lst:critic prompt},captionpos=b]
System: You are a prompt generator designed to generate textual description inputs for activities as a Python dictionary. Do not provide anything other than a prompt.   
User: Generate a dictionary of 25 descriptions for each activity in the list of activities =  [ "Walking", "Running", ...] }. 
\end{lstlisting}
\label{fig:prompt}
\end{figure}

The IMU Feature Extractor (IMU-FE) processes raw multivariate IMU signals into compact, low-dimensional feature sequences, effectively capturing temporal dynamics and reducing noise. These extracted features are then passed to the IMU Encoder, which transforms them into latent embeddings that encapsulate the essential characteristics of the activity patterns. This transformation into a latent space facilitates more effective alignment with the textual modality by providing a structured representation that is both discriminative and semantically meaningful. On the text side, we utilize a frozen pretrained CLIP text encoder~\cite{radford2021learning} to generate embeddings for activity descriptions, which are crafted using prompt engineering techniques with GPT-4, shown in Listing~\ref{lst:critic prompt}.



This encoder ensures that the semantic richness of natural language labels is preserved and accurately represented. Finally, two modality-specific projection heads,  one for IMU embeddings and one for text embeddings, map their respective representations into a common embedding space. This shared space enables direct comparison and similarity computation between sensor data and text descriptions, forming the basis for our contrastive learning objective. This design allows IMU-CLIP to effectively learn cross-modal relationships crucial for few-shot activity classification and privacy-aware sensing applications.

{\bf Training.} To train IMU-CLIP, we employ a supervised contrastive loss $\mathcal{L}^{\text{sup}}$ as defined in~\cite{khosla2020supervised}.   This loss leverages labeled IMU data to train the model in a supervised manner by simultaneously considering multiple positive and negative pairs within a batch. Specifically, for each anchor sample, the objective is to pull the embeddings of all positive samples (i.e., those sharing the same class label) closer in the embedding space, while pushing the embeddings of negative samples (from different classes) further apart. Formally, given a batch of N samples, the supervised contrastive loss is defined as:

\begin{equation}
    \mathcal{L}^{sup} =  \sum_{n \in \mathcal{N}}  -\frac{1}{|P(n)|} \sum_{p \in P(n)} \log  \frac{\exp \left(\left( z_n \cdot {z_p}\right)/{\tau} \right)}{\sum_{a \in A(n)} \exp \left( \left( z_n \cdot {z_a} \right)/{\tau} \right)} 
    \label{equation:supcon}
\end{equation}
Here, $P(n)$ denotes the set of indices of all positive samples in the batch corresponding to the anchor $n$, while $A(n) \equiv N \setminus \{n\}$ represents the set of all indices in the batch excluding the anchor itself. The vectors $z_n$ and $z_p$ are the normalized embeddings of the anchor and positive samples, respectively, and $\tau$ is a temperature hyperparameter controlling the concentration of the distribution. This loss encourages the model to cluster embeddings of samples from the same class tightly together while pushing embeddings from different classes farther apart in the shared embedding space.

After completing the training process, IMU-CLIP learns to identify the similarity mapping between the IMU sensor data embeddings and the corresponding textual descriptions for the given classes.
Similarity score is computed using the dot products of the given IMU data projected into $C$ dimensions and the textual embeddings of classes in the list of $C$ textual classes. Let $I$ be the IMU embeddings and $T_c$ be the textual embedding of a class $c$, then the similarity score associated with class $c$ can be defined as
\begin{equation}
s_c =  I \cdot {T_c}
\end{equation}
To improve performance, we augment the input text description with a large set of activity descriptions generated using OpenAI's GPT4.0.
We use the text encoder from OpenCLIP, which is pre-trained with a large amount of publicly available text and capable of semantically generating sequences of tokens.
By adding time-series sensor data and a corresponding large amount of activity descriptions in natural language form, generated by carefully crafted prompts, we make the model transfer its capability to perform IMU sensor data classification in a zero-shot manner. The classification performance can be further improved by supplying a few shots (samples) of unseen classes previously considered in the zero-shot technique.

\subsection {Privacy Personalizer}
The Privacy Personalizer is a client-side module that enables users to dynamically define and manage their privacy preferences concerning activity inferences from sensor data. This module allows individuals to explicitly control which inferences are considered sensitive and should be obfuscated, and which are deemed acceptable for disclosure or use by third-party applications.
Users specify their preferences through a simple interface, classifying activity labels into three distinct categories, adapted from the RAE privacy taxonomy~\cite{malekzadeh2018replacement}:
\begin{itemize}
\item \textit{White-listed classes} (W): These are non-sensitive activities that users are comfortable sharing. For example, "walking" may be a white-listed activity that contributes to benign fitness tracking applications.

\item \textit{Black-listed classes} (B): These are highly sensitive activities that users do not wish to be inferred or disclosed under any circumstances. For instance, "smoking" may be black-listed due to personal, health, or insurance-related privacy concerns.
\item \textit{Gray-listed classes} (G): These activities are considered neutral --- users do not object to their disclosure, and service providers typically do not find them relevant. An example might be "standing," which does not carry strong privacy implications in most contexts.
This classification is dynamic, allowing users to update their preferences in real time based on context, location, or changes in sensitivity. 
\end{itemize}
Once preferences are specified, the Privacy Personalizer communicates the current privacy configuration to downstream modules, such as the ACT-SANITIZER, which uses this input to selectively transform or replace sensitive activity traces while preserving the utility of non-sensitive data. 



\subsection{ACT-SANITIZER} 

Once the IMU-CLIP model is trained and deployed, the next phase in the PrivCLIP pipeline focuses on safeguarding user privacy by replacing sensor data associated with black-listed activities. This is accomplished through two key components: the Activity Description Sanitizer and an IMU data synthesizer named IMU-GPT. These modules work in tandem to transform sensitive sensor readings into representative sequences of non-sensitive, gray-listed activities. The complete transformation pipeline is illustrated in Figure~\ref{fig:CLIP-HAR-Replacement}.

\begin{algorithm}[t]
\footnotesize
\caption{PrivCLIP Replacement Algorithm.\\
Input: $X$ is the raw sensor data; $B$ is the set of black-label activities, $G$ is the set of gray-list activities, and $W$ is the set of white-list activities. $\textit{act-description}$ is the list of activity descriptions and $I$ is the trained IMU-CLIP model.\\
Output: $X^{'}$ is the transformed sensor data after the sanitization.}
\label{alg:transform_imu}
\begin{algorithmic}[1]

\Procedure{PrivCLIP}{$X$}

    \State $ \textit{top-K-activities} \gets \Call{IMU-CLIP}{X, B}$ \Comment{few-shot detection} 

    \If{$  \textit{top-K-activities}(1) \in B $} \Comment{If this is a black-listed activity}
    
        \For {$\text{predictions } k \gets \textit{top-K-activities}(i), i = {  2,3,\cdots K} $}
            \If{$k \notin B \And k \in G$}
                \State $A \gets \Call{ACT-Sanitizer}{\textit{act-description(k)}}$
                \State $X^{'} \gets \Call{IMU-GPT}{Act, X}$
                \State \textbf{return} $X^{'}$
            \EndIf
        \EndFor
    \EndIf
    \State \textbf{return} $X$
\EndProcedure

\Function{IMU-CLIP}{$X , B$}
    \State $ \textit{top-K-activities} \gets I(X,B)$ \Comment{compute top-k similarities}
    \State  \textbf{return} $\textit{top-K-activities}$ 
\EndFunction

\Function{ACT-Sanitizer}{$g$}
    \State  \textbf{return} $Act$ \Comment{generate non-sensitive Activity description using GPT4.0}
\EndFunction

\Function{IMU-GPT}{$A, X $}
    \State  \textbf{return} $X^{'}$ \Comment{generate non-sensitive IMU from text description using IMU-GPT}
\EndFunction

\end{algorithmic}
\end{algorithm}

Algorithm~\ref{alg:transform_imu} outlines the user-controllable privacy transformation implemented by PrivCLIP.
Initially, PrivCLIP performs activity detection using the trained IMU-CLIP model in a few-shot learning setup. It classifies each incoming IMU sensor sequence by computing its similarity to a set of predefined textual activity descriptions, which are curated to include activities from the user's white-listed ($W$), black-listed ($B$), and gray-listed ($G$) preference sets.

Formally, for a given input IMU signal $x \in \mathcal{X}$, IMU-CLIP calculates similarity scores $s_i$ between the encoded embedding of $x$ and the embeddings of the textual descriptions ${t_i}$ corresponding to the user's defined activity set $W \cup B \cup G$. The top-K activities are selected based on the highest similarity values, denoted as:

\begin{equation}
 TopKActivities(x) = \left\{t_i| s_i \in \text{Top-K} \Big(\{s_j\}_{j=1}^{|W \cup B \cup G|}\Big) \right\}    
\end{equation}

We then pass the \textit{TopKActivities} to the ACT-SANITIZER module, which operates in conjunction with the privacy personalizer. As described earlier, the privacy personalizer enables users to dynamically specify their privacy preferences by categorizing activities into white-listed ($W$), black-listed ($B$), and gray-listed ($G$) sets. These user-defined categories guide the transformation process carried out by the ACT-SANITIZER. 

The ACT-SANITIZER executes the PrivCLIP transformation algorithm as follows: if the top-ranked activity in the \textit{TopKActivities} belongs to the black-listed set $B$, the algorithm searches for the next most similar activity within the \textit{TopKActivities} that belongs to the gray-listed set $G$. This ensures that the replacement activity remains semantically and statistically similar to the original, minimizing deviation in feature representation and thus preserving utility while enforcing privacy.
Once a suitable gray-listed replacement activity is selected, a textual description for it is generated using the prompt engineering techniques described in the previous section. This description is then fed into IMU-GPT~\cite{leng2024imugpt}, a human motion generation framework that synthesizes realistic time-series sensor signals from natural language activity descriptions. Leveraging a state-of-the-art motion synthesis model, IMU-GPT produces a sanitized version of the sensor data representing the non-sensitive activity. This process results in a privacy-preserving transformation of the original sensor data, effectively masking sensitive activity inferences while preserving the overall utility of the data.

\section{Experimental setup}


\subsection{Dataset}

\begin{table*}[t]
\centering
\small
\begin{tabular}{|c|c|c|c|c|}
\toprule
Dataset      & \begin{tabular}[c]{@{}c@{}}Subject\\ count\end{tabular} & \begin{tabular}[c]{@{}c@{}}Sensors \\ used\end{tabular} & \begin{tabular}[c]{@{}c@{}}Number of\\ classes\end{tabular} & \begin{tabular}[c]{@{}c@{}}Number of\\ features\end{tabular} \\ \midrule
Skoda & 1  & 3D accel. & 10 & 54                                                            \\ \hline
Opportunity & 4  & accel., gyro. & 17 & 30                                                           \\ \hline
Hand-gesture & 2 & accel., gyro. & 11 & 15                                                            \\ \bottomrule
\end{tabular}
\caption{Summary of datasets used in our experiments.}
\label{tab:har_dataset}
\end{table*}

We conduct our experiments on three human activity recognition benchmark datasets, shown in Table~\ref{tab:har_dataset}. 

\textbf{Skoda dataset~\cite{zappi2008activity}} comprises 11 activities performed by assembly-line workers in a car production environment by a subject wearing 19 3D accelerometers on both arms and performing a set of experiments using sensors placed on the two arms of a tester.


\textbf{ Opportunity dataset~\cite{chavarriaga2013opportunity}} is a benchmark dataset for HAR that contains daily life human activities performed by four subjects. The data comprises 113 sensory readings, and there are 18 gesture classes.

\textbf{Hand-gesture dataset~\cite{bulling2014tutorial}} consists of 11 hand gestures recorded from body-worn accelerometers and gyroscopes of two subjects repeating all activities for 26 times.

All datasets are normalized with zero mean and unit standard deviation.
In all datasets, we use 80\% of time windows for the training phase, and the remaining 20\% is used for the tests. 
We randomly select all available samples of non-sensitive activities and only $k$ samples of sensitive activities from the training dataset, where $k$ corresponds to the number of shots in the experiment (referred to as \textit{k-shot}). Unless otherwise specified, $k$ is set to 64 in our experiments.

\subsection{Baseline techniques}
We use the following two baseline techniques to evaluate our technique quantitatively:

\textbf {Replacement autoencoder (RAE)~\cite{malekzadeh2018replacement}} is based on an autoencoder that learns to replace sensitive activity sensors with non-sensitive sensor readings based on a predefined replacement mapping.
This is achieved by training the RAE to output gray-list activities if sensor data corresponding to a black-list sensor activity is given. In the case of other classes of sensor data, the RAE reconstructs the sensor data corresponding to the same activity class. The reconstruction loss is computed using the reconstructed sensor data and the randomly chosen replaced data that belong to non-sensitive data according to the replacement mapping defined by the user.


\textbf {Few-shot HAR (FS-HAR) ~\cite{10.1007/978-981-96-2468-3_16}} is a few-shot HAR framework incorporating a feature extractor and a set of autoencoders.
The output features of the feature extractor are employed as input for training the autoencoders within the framework. During training, the autoencoders learn to reconstruct the given input feature.
The first autoencoder is trained with data belonging to base class activities with many samples, and other autoencoders are trained with one kind of new class or classes with a few samples of data.
If the first autoencoder is fed data from a class it was not trained with, the reconstruction loss will be high and considered as a new or unseen class. This will be sent to another set of autoencoders in the framework, each trained with a few shots of new classes. Based on the correlation between the input and output of a set of autoencoders, a similarity score is computed as mentioned in the paper ~\cite{10.1007/978-981-96-2468-3_16}. The class with high similarity beyond a set threshold is then assigned to that input data. 
We then use the same replacement Algorithm~\ref{alg:transform_imu} to replace the black-list activity with the next similar gray-list activity.



\subsection{Model}

We have used multiple models to implement our frameworks and the baseline techniques.

\textbf{IMU-CLIP} architecture has an IMU and a text encoder. The text encoder is a pre-trained frozen text encoder from the open-source implementation of OpenAI’s CLIP - OpenCLIP, using a backbone network of a ViT-B/32.
For the IMU encoder, we adopt a vision transformer-based model that processes time-series data by splitting it into patches, analogous to token sequences in NLP. The IMU encoder comprises a 2D convolutional layer, three self-attention layers, and a dense layer with ReLU activation. Both the IMU and text projection heads are implemented as linear layers projecting embeddings into a shared 512-dimensional space. The model is trained using the AdamW optimizer with a supervised contrastive loss~\cite{khosla2020supervised}. We set the learning rate to 0.001 for the IMU encoder, IMU projection head, and text projection head, and a lower learning rate of 0.0001 for the frozen text encoder. The entire framework is implemented in PyTorch and trained for 200 epochs with a batch size of 32.


\textbf{Activity Classifier} is based on a Convolutional Neural Network (CNN)-based network with three layers, with ReLU activation, followed by two fully connected dense layers. 
We use Adam optimizer with a learning rate of 0.001 and a loss function of categorical cross-entropy.
The activity classifier is implemented using the Keras framework, and we train it for 200 epochs with a batch size of 64.

\textbf{RAE} is an autoencoder structure with an input layer and five hidden layers with SeLU activation. All experiments are performed over 30 epochs, with a batch size of 128. The loss function used is Mean Squared Error (MSE).
The model is implemented in the Keras framework.

\textbf{FS-HAR} is a few-shot HAR framework comprising a feature extractor and a set of autoencoders.
The feature extractor in the framework consists of 2 CNN layers with nodes of 64 and 32, followed by a dense layer, all with ReLu activations. We use Adam optimizer, and the model is trained for 500 epochs with a learning rate of 0.0005.
The autoencoders have three fully connected layers with ReLu activation for both the encoder and decoder networks. The model is trained with the Adam optimizer and a learning rate of 0.001. MSE is the loss function.
The model is implemented in the Keras framework.


\section{Evaluation}

In this section, we evaluate the performance of our few-shot PrivCLIP in terms of few-shot detection and replacement in various experimental settings.
\begin{table*}[t]
\centering

\begin{tabu}{|l|c|c|c|c|} 
\toprule
\multicolumn{1}{|c|}{Dataset} & \begin{tabular}[c]{@{}c@{}}Training time mapping\\ $\{black\-list\} \rightarrow \{gray\-list\}$\end{tabular} & \begin{tabular}[c]{@{}c@{}}Inference time mapping\\ $\{black\-list\} \rightarrow \{gray\-list\}$ \end{tabular} & \begin{tabular}[c]{@{}c@{}}RAE\\(F1)\end{tabular} & \begin{tabular}[c]{@{}c@{}}PrivCLIP \\(F1)\end{tabular} \\ 
\midrule
Skoda & $(\{1,5,6,7\} \rightarrow \{0,2,3\})$ & $\{1,5,6,7\}\rightarrow \{4,8,9\}$ & 0.01 & \textbf{0.94} \\ 
\hline
Opportunity & $\{1,2,3,4,5,6,7,8\} \rightarrow \{0\}$ & $\{1,2,3,4,5,6,7,8\} \rightarrow \{1\}$ & 0.03 & \textbf{0.85} \\ 
\hline
Hand-gesture & $\{5,6,7,8\} \rightarrow \{0\}$ & $\{5,6,7,8\} \rightarrow \{1\}$ & 0.04 & \textbf{0.88} \\
\bottomrule
\end{tabu}
\caption{Performance comparison on activity replacement in a dynamic privacy setting on various datasets.}
\label{tab:unseen_detection_dynamic}
\end{table*}

\subsection{Dynamic privacy scenario}

We begin by evaluating the scenario where users choose to update their privacy preferences after the model has been deployed on their device.
To investigate this, we compare the performance of PrivCLIP and RAE  in a dynamic privacy scenario.
In the case of PrivCLIP, the model is trained with 64 samples from predefined privacy classes or black labels \{1,5,6,7\}, and during the run time, the user dynamically chooses their gray labels to replace with.
%
%
%
%
%
%
%
%
Similarly, in the case of RAE, the model is trained using a fixed set of black labels \{1,5,6,7\} and a fixed set of gray labels \{0,2,3\}. Table~\ref{tab:unseen_detection_dynamic} shows that RAE produces low replacement performance when we dynamically change the gray label to \{4,8,9\}. 
For instance, with the Skoda dataset, RAE's replacement classification performance is very poor, 0.01, while CLP-HAR adapts to dynamic privacy needs and replaces with a high F1 Score of 0.94.
This is because RAE is not trained to learn the replacement strategy dynamically post-training.
In the case of PrivCLIP, replacement performance is high, as indicated by the high F1-score.
This proves that PrivCLIP doesn't need prior privacy annotation for each class during the training phase, as described in the design section. In other words, privacy classification, such as black-label, gray-label, and white-label classification, can be done dynamically by the user post-deployment. In contrast, RAE requires retraining and redeployment as the privacy requirement changes. 
%
%
%
PrivCLIP offers dynamic privacy controllability and can achieve2 better results than RAE in all combinations.

\subsection{Performance comparison}
Next, we evaluate the performance of our few-shot privacy-preserving sensing framework on benchmark datasets against other baseline techniques. 
As specified in the above section, we assign activities to three categories according to user privacy preferences: black, gray, and white labels.
%

\begin{table*}[t]
\centering

\begin{tabu}{|l|l|c|c|c|c|c|c|} 
\toprule
\multicolumn{1}{|c|}{\multirow{3}{*}{Dataset}} & \multicolumn{1}{c|}{\multirow{3}{*}{Privacy classification}} & \multirow{3}{*}{\begin{tabular}[c]{@{}c@{}}Original data Before\\transformation (F1)\end{tabular}} & \multicolumn{5}{c|}{\begin{tabular}[c]{@{}c@{}}After transformation \\(F1)\end{tabular}} \\ 
\cline{4-8}
\multicolumn{1}{|c|}{} & \multicolumn{1}{c|}{} &  & \multirow{2}{*}{RAE} & \multirow{2}{*}{FS-HAR (64-shot)} & \multicolumn{3}{c|}{PrivCLIP} \\ 
\cline{6-8}
\multicolumn{1}{|c|}{} & \multicolumn{1}{c|}{} &  &  &  & 8-shot~ & 32-shot~ & 64-shot \\ 
\midrule
\multirow{3}{*}{Skoda-left} & W: \{4,8,9,10\}) & 98.27 & 90.64 & 81.34 & 90.61 & 95.12 & 96.12 \\ 
\cline{2-8}
 & B: \{1,5,6,7)\} & 97.56 & 0.08 & 12.31 & 9.12 & 4.80 & 0.17 \\ 
\cline{2-8}
 & G: \{0,2,3\} & 95.98 & 89.51 & 78.91 & 95.13 & 95.69 & 96.23 \\ 
\hline
\multirow{3}{*}{Opportunity} & W: \{9,10,11,12,13,14,15,16,17\} & 94.58 & 74.76 & 72.34 & 75.56 & 79.12 & 82.85 \\ 
\cline{2-8}
 & B: \{1,2,3,4,5,6,7,8\} & 88.79 & 2.19 & 13.43 & 2.18 & 1.17 & 0.95 \\ 
\cline{2-8}
 & G: \{0\} & 91.42 & 86.95 & 83.37 & 87.01 & 87.43 & 89.01 \\ 
\hline
\multirow{3}{*}{Hand--Gesture} & W: \{1,2,3,4,9,10,11\} & 93.65 & 72.66 & 70.65 & 73.01 & 74.45 & 74.62 \\ 
\cline{2-8}
 & B: \{5,6,7,8\} & 94.17 & 0.45 & 15.45 & 0.51 & 0.46 & 0.44 \\ 
\cline{2-8}
 & G: \{0\} & 95.66 & 96.89 & 78.81 & 96.93 & 97.04 & 97.11 \\
\bottomrule
\end{tabu}
\caption{Comparison with Baseline techniques. Private activity classification after transformation.}
\label{tab:RAE_comparison}
\end{table*}
The results are plotted in Table~\ref{tab:RAE_comparison}, where the first column contains the dataset and various combinations of sensitive, non-sensitive, and desired classes used in the experiment. 
In the second column, we provide the model performance of the activity classification task on the original data. As seen in the table, the F1-score on the original data before transformation is high across different datasets and combinations of classes.
We then present the next columns with the performance of various techniques used for sensor data transformation.

\begin{figure}[t]
    \centering
    \includegraphics[width=3.2in]{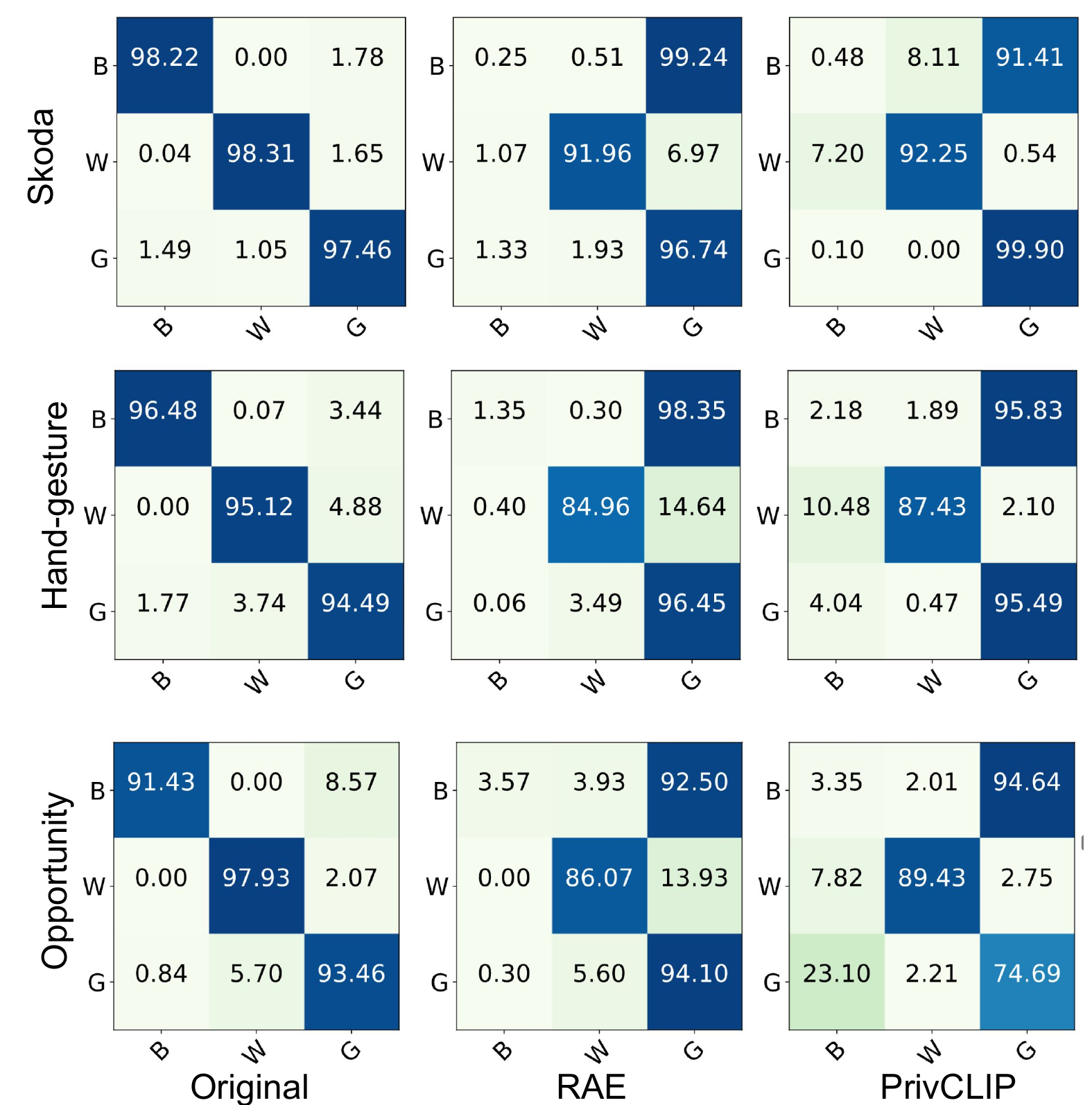}
    \caption{Classification performance comparison. The X-axis is the predicted label, and the Y-axis is the true label.} 
    \label{fig:3x3_transformed}
\end{figure}

In the case of RAE, we use a complete training dataset to train the replacement autoencoder, and the performance on test data is shown. While FS-HAR's base autoencoder is trained with gray and white data classes, the other set of new-class autoencoders is trained with 64 data samples each from classes belonging to black-list classes.
In the case of PrivCLIP, we consider three settings of k-shot learning, where k is the number of samples of each new class used in the training set, and all training samples from the gray and white-list classes.
We can see that PrivCLIP outperforms FS-HAR in all scenarios and performs better than RAE in most of them. Recall that RAE is trained with all the data, but the other techniques are trained with a few shots from black-label classes.
We also show the confusion matrix for the classification performance before and after sensor data transformation using RAE and privCLIP in Figure~\ref{fig:3x3_transformed}. 

PrivCLIP preserves user privacy by transforming the sensitive activity with non-sensitive activity in a few-shot learning. The performance improves as we increase the size of k or the number of training samples available during the training phase. 

\subsection{Few-shot activity detection}

\begin{figure}[t]
    \centering
    \subfigure[Class-wise on IMU-CLIP]{\includegraphics[width=1.64in]{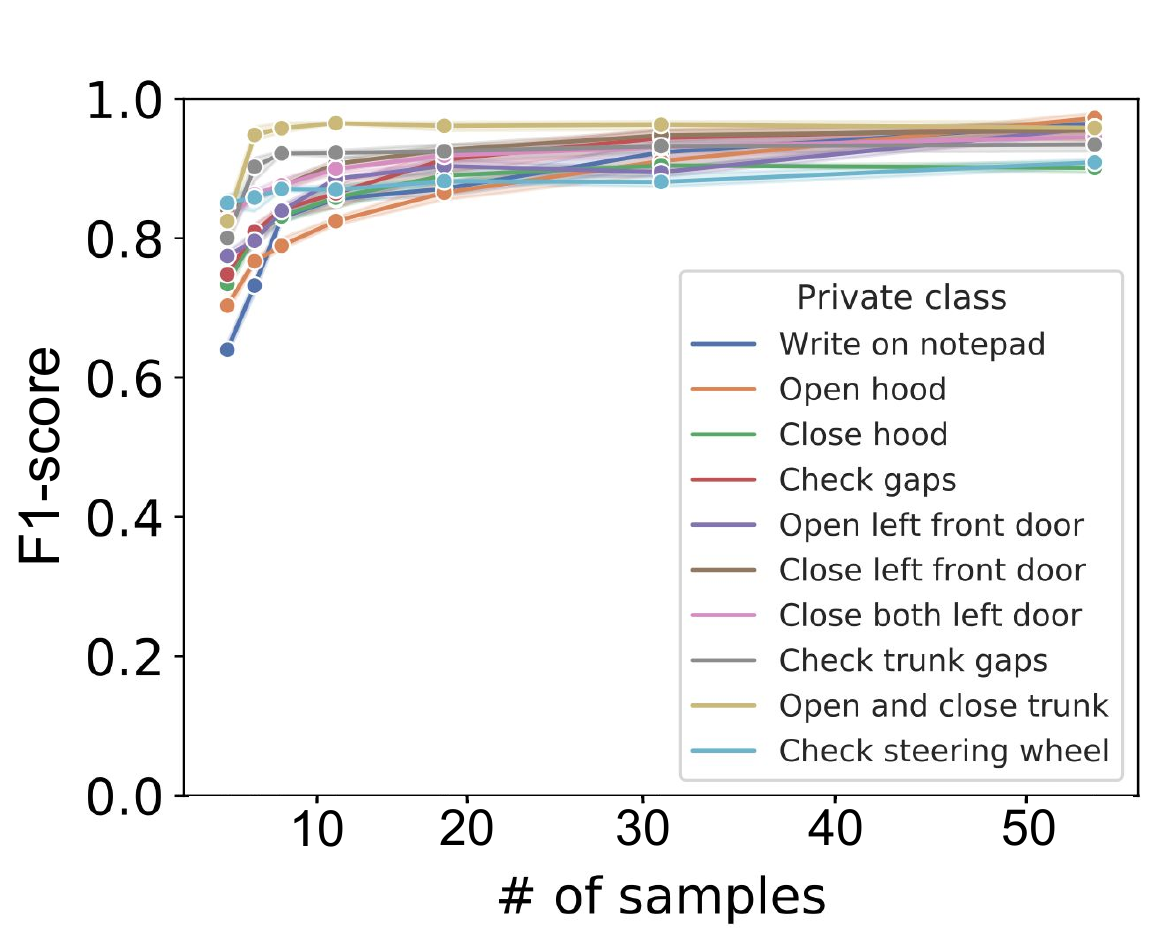}}
    \subfigure[IMU-CLIP and FS-HAR]{\includegraphics[width=1.64in]{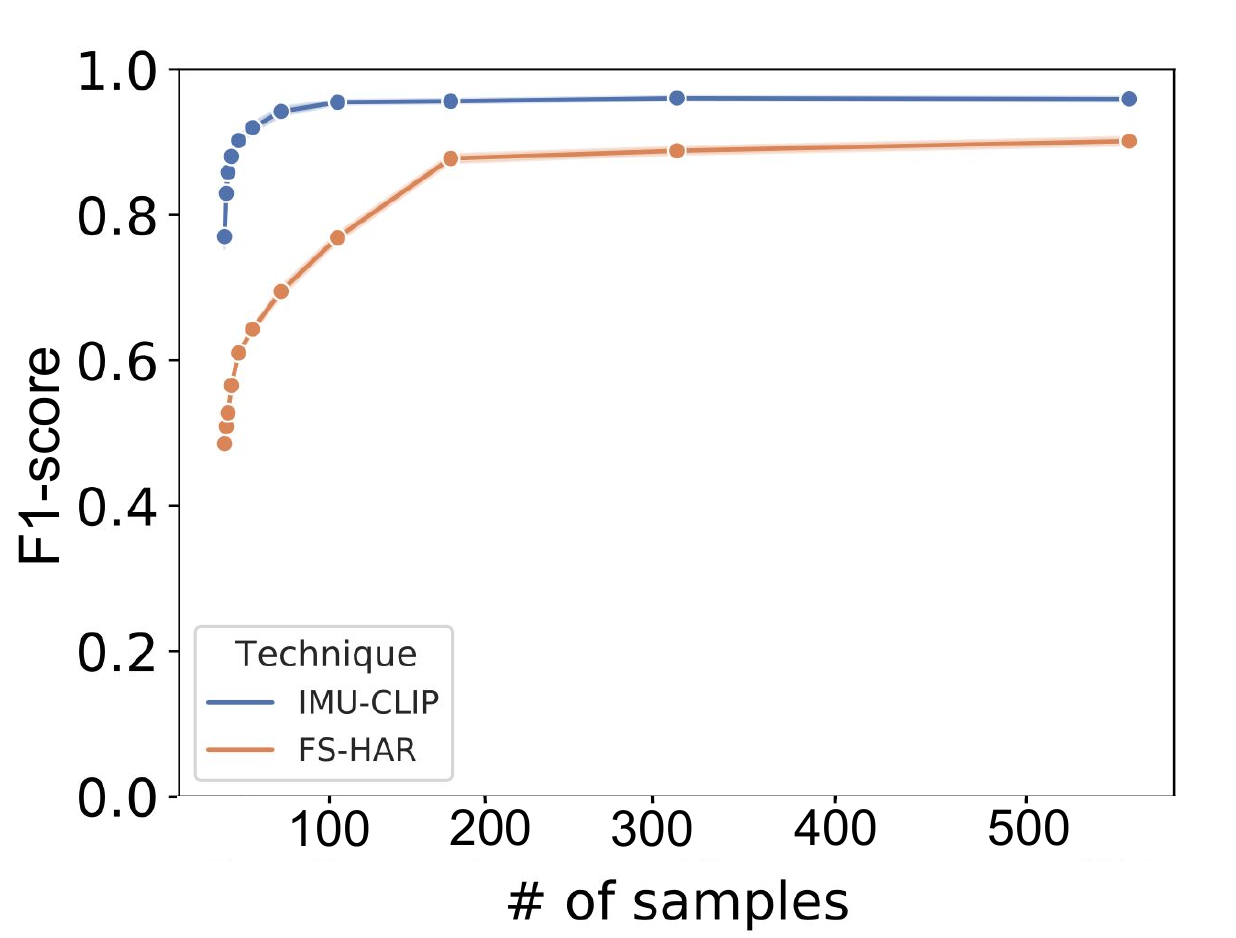}}
    \caption{Performance comparison of few-shot activity detection techniques on the Skoda dataset.}
    \label{fig:fs-perfm}
\end{figure}

\begin{figure}[t]
    \centering
    \subfigure[Skoda dataset]{\includegraphics[width=1.in]{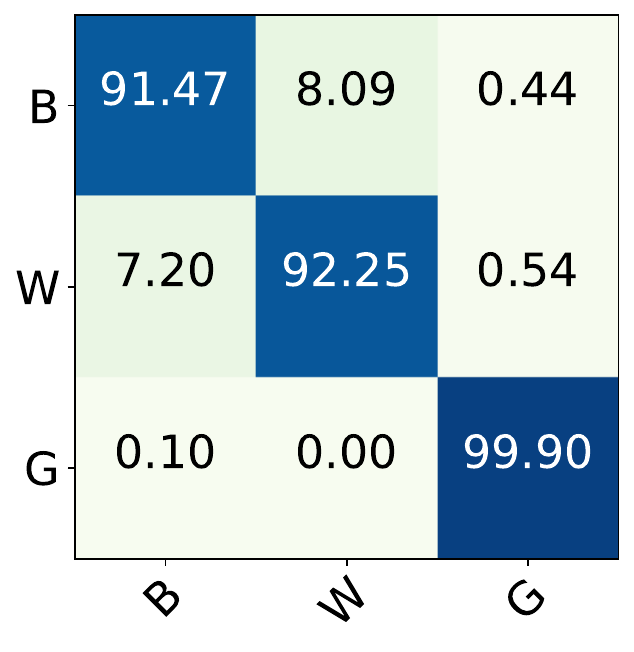}} 
    \subfigure[Hand-gesture dataset]{\includegraphics[width=1.in]{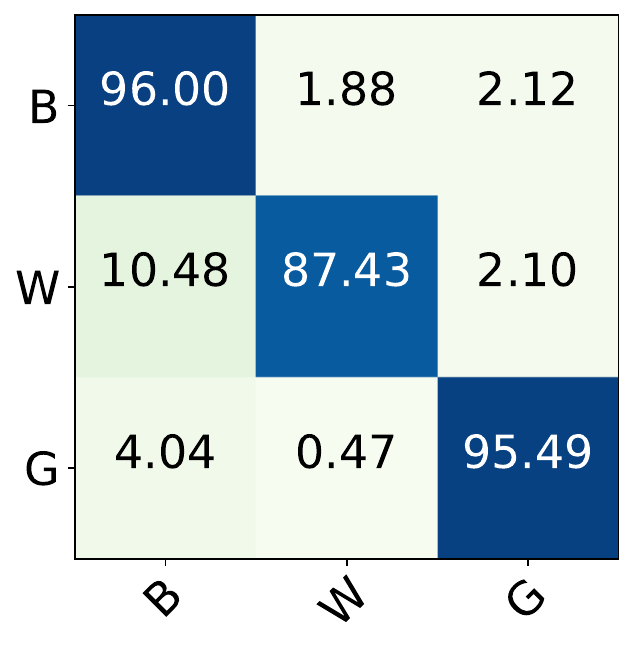}} 
    \subfigure[Opportunity dataset]{\includegraphics[width=1.in]{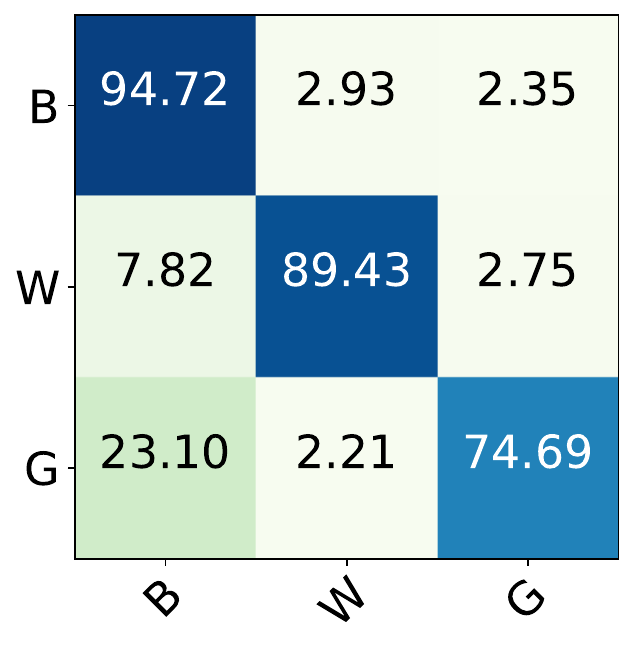}} 
    \caption{Few-shot detection with shot size of $k = 64$ of sensitive activities. 
    }
    \label{fig:PrivCLIP-Detection}
\end{figure}

In this experiment, we fix a few-shot sensitive classes and vary the number of samples available for training by randomly selecting from the given dataset. We start with as few as no samples (zero-shot), then one sample, and increase it exponentially. As seen in the Figure ~\ref{fig:fs-perfm}(a),  the detection performance improves as we increase the number of sensitive samples (few-shot). We do the same experiments across all classes in the dataset.
As seen, with a minimal number of samples between 4 and 8, IMU-CLIP can more accurately detect the activities. Specifically, with a few shots of 4, all classes can be detected with an F1 score above 0.8, which is 0.94 in the case of activity—\textit {open and close trunk} in the Skoda dataset.
A confusion matrix for few-shot detection is shown in Figure~\ref{fig:PrivCLIP-Detection}.

We further evaluate the performance of IMU-CLIP in comparison to the baseline technique, FS-HAR, specifically regarding few-shot detection. In this assessment, we analyze the few-shot detection capabilities of PrivCLIP alongside the autoencoder-based FS-HAR technique. For this experiment, we calculate the F1-score for few-shot detection by treating one class of data at a time as a new or rarely seen class, while considering all other classes as the base or seen classes. Finally, we compute the average of all F1-scores for each technique.
The results of comparing FS-HAR and IMU-CLIP for few-shot detection using the Skoda dataset are presented in Figure~\ref{fig:fs-perfm}(b). For FS-HAR, the average F1-score in a zero-shot scenario is approximately 0.5, while IMU-CLIP achieves an average zero-shot detection score of around 0.7. As the number of samples increases, both techniques show improved performance, but IMU-CLIP demonstrates a more pronounced increase with fewer samples. For example, with 128 samples, FS-HAR reaches an F1-score of 0.88, whereas IMU-CLIP achieves an F1-score of 0.96.


Next, we compare the performance of IMU-CLIP in zero-shot and few-shot detection modes. Table~\ref{tab:unseen_detection_table} shows the mean F1-score and standard deviation of private activity prediction in a zero-shot, one-shot, and few-shot setting with four samples across all classes in the dataset, taking one private class at a time. Across all datasets, the performance increases from 12\% to 19\%. We also see a significant increase in detection performance with one shot.

\begin{table}[t]
\centering

\begin{tabu}{|l|c|c|c|} 
\toprule
Dataset & Zero-shot &  One-shot & Few-shot (4 samples) \\ 
\midrule
SKODA  & 0.77 $\pm 0.05$  & 0.84 $\pm 0.04$ & 0.85 $\pm 0.02$ \\
\hline
Opportunity &  0.76 $\pm 0.04$ & 0.86 $\pm 0.03$  & 0.89 $\pm 0.02$   \\
\hline
Hand gesture&  0.72 $\pm 0.04$  &0.88 $\pm 0.04$ &  0.91 $\pm 0.03$ \\
\bottomrule
\end{tabu}
\caption{Zero-shot and Few-shot private activity detection performance (F1-score) of PrivCLIP on various datasets.}
\label{tab:unseen_detection_table}
\end{table}




\begin{figure}[t]
    \centering
    \subfigure[Label]{\includegraphics[width=1.69in]{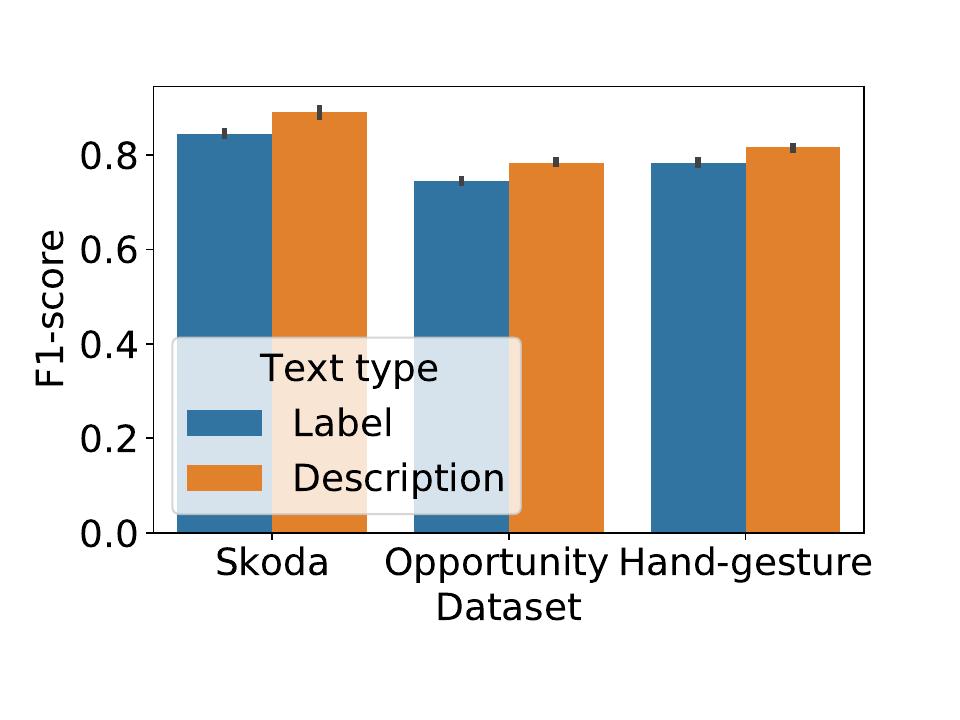}}
    \hspace{-0.5cm}
    \subfigure[Description]
    {\includegraphics[width=1.69in]{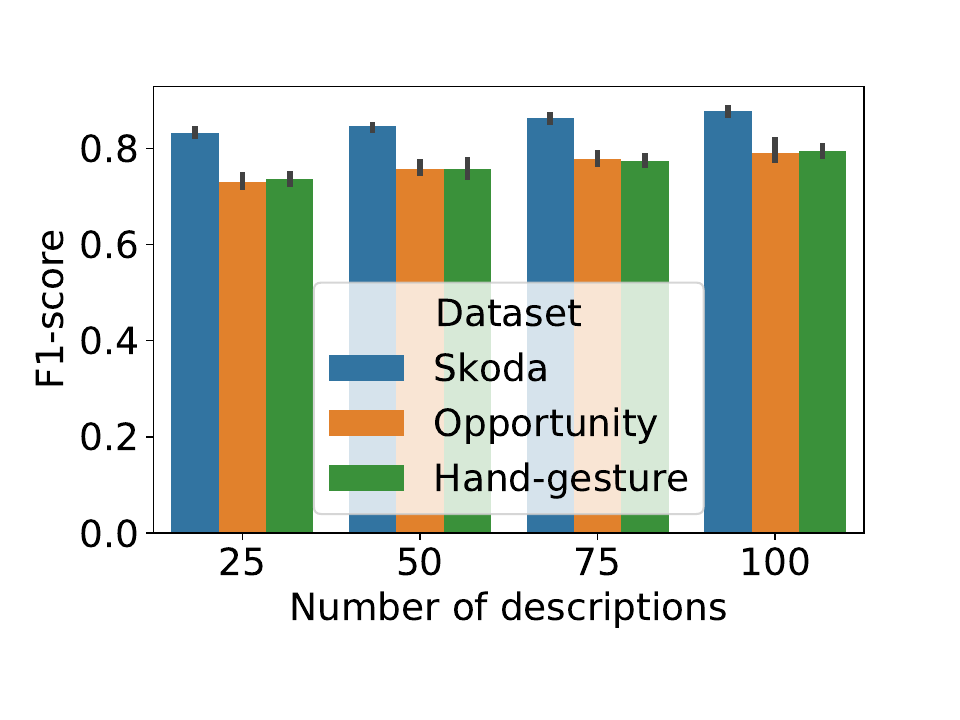}}
    \caption{Effect of varying labels and descriptions.}
    \label{fig:desc}
\end{figure}



IMU-CLIP can detect activities in a few-shot manner and provide higher accuracy compared to other baseline techniques, such as FS-HAR, which has a very low number of samples . Even with no samples, IMU-CLIP achieves close to 70\% accuracy and improves with adding a few samples.

\subsection{Effect of activity description}
During the IMU-CLIP training process, in addition to the activity class label, we include textual descriptions of the activity class to which each activity belongs. We use the GPT4.0 model to generate 25 activity descriptions for each activity in all experiments unless specified otherwise. Similarly, we use GPT4.0 activity descriptions during the detection phase instead of the activity name. We fix a sample size of 64 in all experiments. We compare the performance of text input given as the class label with that of the description.
As shown in Figure~\ref{fig:desc}(a), IMU-CLIP achieves improved performance of up to 6\% when using textual description instead of class names using prompt engineering during few-shot detection.





By providing activity descriptions instead of short class names, the IMU-CLIP technique based on LLM-based models can perform better since it is pretrained with a large amount of natural language texts and helps identify the similarity and dissimilarity within the data semantically. 

\subsection{Effect of number of descriptions}
To compare the impact of the quantity of textual descriptions used for each activity class on detection performance, we vary the number of activity descriptions for each class to 25,50,75, and 100 descriptions, and the results are shown in Figure~\ref{fig:desc}(b).
We fix a k-shot with k = 8 for all datasets in this experiment.
The few-shot activity detection performance calculated in terms of F1-score increases as the number of activity descriptions associated with each activity class in the dataset increases, and this behavior is consistent across all datasets.
\section{Related work}
IoT devices are ubiquitous, and the large amount of data they generate can help enable various data-driven applications from home automation and health monitoring to powering critical applications in safety and security domains ~\cite{hassija2019survey, chathoth2025pcap, melnyk2025hardware}. 
Human activity recognition has become integral to most modern wearable, mobile, and gaming devices~\cite{chathoth2024dynamic}. While these features add insights into healthy living, entertainment, etc., they pose a severe threat to user privacy~\cite{malekzadeh2018protecting}. In wearable devices, the devices share sensor data corresponding to activities with cloud-based services for better analysis and providing data-enabled services. This information is usually sent to a cloud-based classifier. If the cloud service is untrusted, it may be able to infer sensitive attributes and recognize sensitive activities~\cite{malekzadeh2018replacement, malekzadeh2019mobile}.
While data sharing is essential for cloud-based services to leverage AI and ML-based tools to enhance the quality of the application, user-sensitive attribute inference is a significant concern that hinders such data sharing in a data-driven system.
Some techniques, such as differential privacy, homomorphic encryption, etc., are proposed, but each has limitations. While DP claims to provide a privacy guarantee at the expense of utility, more granular privacy controls by the user are often ignored in the DP-based proposed solutions. DP perturbs the data the user shares, but our goal is to share the data required for the utility while preventing the attacker from inferring the sensitive attributes from the given data.

Sensor data transformation is recognized as a technique for preserving privacy~\cite{malekzadeh2018protecting, malekzadeh2018replacement, malekzadeh2019mobile, raval2019olympus}. RAE is an autoencoder-based technique that first learns a static transformation mapping from sensitive data to nonsensitive data and then replaces discriminative features that correspond to sensitive inferences with features more commonly observed in the nonsensitive inferences~\cite{malekzadeh2018replacement}. This approach only works when the sensitive and nonsensitive classes are predefined before model training, which limits its ability for dynamic privacy control. 
While there are works on user control over privacy, none are towards human activity inference privacy~\cite{asif2018increasing, chhetri2022user}.
Our approach aims to provide dynamic user control over their privacy preferences without retraining or redeployment of the model.
Additionally, RAE requires a large amount of data corresponding to the sensitive classes, which is impractical, thereby introducing the need for a few-shot learning technique in the sensor domain. Several studies focus on few-shot learning based human activity recognition~\cite{10.1007/978-981-96-2468-3_16, ganesha2024few, feng2019few, wanyan2024comprehensive, xue2024leveraging, al2020zero, tong2021zero, cheng2013towards, nag2023semantics, zhang2020zstad, zhang2022tn, deng2020few}. 

FS-HAR is a few-shot HAR detection framework based on a deep feature extractor and a set of autoencoders that learn to identify few-shot classes in an unsupervised setting based on similarity score~\cite{10.1007/978-981-96-2468-3_16}. While FS-HAR can carry out few-shot learning, the work does not analyze the performance over the shot size. Moreover, FS-HAR requires knowing sensitive classes' details in advance since it computes the similarity by assigning an autoencoder to each unseen class.
Similarly, a recent work proposes a multi-modality few-shot activity recognition system(FSAR) by augmenting the motion video and action images~\cite{ruan2024advances}.
Another method, ZeroHAR, employs contrastive learning as a zero-shot technique. It enhances motion data by incorporating sensor context features, such as sensor position and sensor type, to align embeddings in a contrastive manner.~\cite {chowdhury2025zerohar}.
TS2ACT is a few-shot HAR based on cross-modal augmentation using text and images alongside sensor data~\cite{xia2024ts2act}.
ADLLLM is an LLM-based technique transforming sensor data into text to perform zero-shot activity recognition~\cite{civitarese2024large}. 
Similarly, Cross-domain HAR is a transfer learning based few-shot human activity recognition framework based on the teacher-student self-training paradigm~\cite{thukral2025cross}.

Unlike prior techniques, we take a different approach by leveraging the power of large language models and contrastive learning techniques to develop few-shot detection skills on activity recognition based on sensor data without compromising the utility.
To provide dynamic data transformation, we generate synthetic data after sanitizing the sensitive parts of motion data according to the user's privacy preferences. While there are techniques to create synthetic IMU sensor signals for activities ~\cite{norgaard2018synthetic, alharbi2020synthetic, leng2024imugpt}, these methods are typically used for training sample generation, not privacy-preserving systems.
Our approach differs from the previously proposed few-shot technique for sensor data transformation. We use contrastive learning-based methods to predict the transformation to a next-best similar nonsensitive activity and an IMU generator to generate sensor data to replace sensitive data while preserving the utility dynamically.


\section{Conclusion}
This paper presents a solution for a utility-aware dynamic privacy-preserving system based on a contrastive learning technique on IMU sensor data. We assess the performance of our technique, PrivCLIP, in detecting sensitive activities using a few-shot learning approach. Additionally, we transform the sensor data related to these sensitive activities into non-sensitive activities dynamically. This method eliminates the need for redeployment or fine-tuning the model for each combination of privacy settings. This is a significant limitation found in prior work, such as the replacement autoencoder that relies on a deterministic mapping of activities and their sensitivity. We demonstrate that our model can identify unseen or rarely encountered sensitive classes across multiple benchmark human activity recognition datasets. We thoroughly compare the performance of PrivCLIP against autoencoder-based few-shot detection and transformation techniques. Our empirical results show that PrivCLIP performs effectively in few-shot detection and replacement of privacy-sensitive activities, without sacrificing the detection of desired activities.

{\bf Acknowledgments.} This research was supported in part by the University of Pittsburgh Center for Research Computing and Data, RRID:SCR\_022735, through the resources provided. Specifically, this work used the H2P cluster, which is supported by NSF award number OAC-2117681.

\bibliographystyle{IEEEtran}

\bibliography{paper.bib}

@article{jain2021differentially,
  title={Differentially private model personalization},
  author={Jain, Prateek and Rush, John and Smith, Adam and Song, Shuang and Guha Thakurta, Abhradeep},
  journal={Advances in neural information processing systems},
  volume={34},
  pages={29723--29735},
  year={2021}
}

@inproceedings{malekzadeh2019mobile,
author = {Malekzadeh, Mohammad and Clegg, Richard G. and Cavallaro, Andrea and Haddadi, Hamed},
title = {Mobile sensor data anonymization},
year = {2019},
isbn = {9781450362832},
publisher = {Association for Computing Machinery},
address = {New York, NY, USA},
url = {https://doi.org/10.1145/3302505.3310068},
doi = {10.1145/3302505.3310068},
booktitle = {Proceedings of the International Conference on Internet of Things Design and Implementation},
pages = {49–58},
numpages = {10},
location = {Montreal, Quebec, Canada},
series = {IoTDI '19}
}

@inproceedings{radford2021learning,
  title={Learning transferable visual models from natural language supervision},
  author={Radford, Alec and Kim, Jong Wook and Hallacy, Chris and Ramesh, Aditya and Goh, Gabriel and Agarwal, Sandhini and Sastry, Girish and Askell, Amanda and Mishkin, Pamela and Clark, Jack and others},
  booktitle={International conference on machine learning},
  pages={8748--8763},
  year={2021},
  organization={PMLR}
}

@article{leng2024imugpt,
  title={Imugpt 2.0: Language-based cross modality transfer for sensor-based human activity recognition},
  author={Leng, Zikang and Bhattacharjee, Amitrajit and Rajasekhar, Hrudhai and Zhang, Lizhe and Bruda, Elizabeth and Kwon, Hyeokhyen and Pl{\"o}tz, Thomas},
  journal={Proceedings of the ACM on Interactive, Mobile, Wearable and Ubiquitous Technologies},
  volume={8},
  number={3},
  pages={1--32},
  year={2024},
  publisher={ACM New York, NY, USA}
}

@article{xue2024leveraging,
  title={Leveraging Foundation Models for Zero-Shot IoT Sensing},
  author={Xue, Dinghao and Fan, Xiaoran and Chen, Tao and Lan, Guohao and Song, Qun},
  journal={arXiv preprint arXiv:2407.19893},
  year={2024}
}

@inproceedings{malekzadeh2018replacement,
  title={Replacement autoencoder: A privacy-preserving algorithm for sensory data analysis},
  author={Malekzadeh, Mohammad and Clegg, Richard G and Haddadi, Hamed},
  booktitle={2018 IEEE/ACM third international conference on internet-of-things design and implementation (iotdi)},
  pages={165--176},
  year={2018},
  organization={IEEE}
}

@article{raval2019olympus,
  title={Olympus: Sensor privacy through utility aware obfuscation},
  author={Raval, Nisarg and Machanavajjhala, Ashwin and Pan, Jerry},
  journal={Proceedings on Privacy Enhancing Technologies},
  year={2019}
}

@article{chavarriaga2013opportunity,
  title={The Opportunity challenge: A benchmark database for on-body sensor-based activity recognition},
  author={Chavarriaga, Ricardo and Sagha, Hesam and Calatroni, Alberto and Digumarti, Sundara Tejaswi and Tr{\"o}ster, Gerhard and Mill{\'a}n, Jos{\'e} del R and Roggen, Daniel},
  journal={Pattern Recognition Letters},
  volume={34},
  number={15},
  pages={2033--2042},
  year={2013},
  publisher={Elsevier}
}

@inproceedings{zappi2008activity,
  title={Activity recognition from on-body sensors: accuracy-power trade-off by dynamic sensor selection},
  author={Zappi, Piero and Lombriser, Clemens and Stiefmeier, Thomas and Farella, Elisabetta and Roggen, Daniel and Benini, Luca and Tr{\"o}ster, Gerhard},
  booktitle={Wireless Sensor Networks: 5th European Conference, EWSN 2008, Bologna, Italy, January 30-February 1, 2008. Proceedings},
  pages={17--33},
  year={2008},
  organization={Springer}
}

@article{bulling2014tutorial,
  title={A tutorial on human activity recognition using body-worn inertial sensors},
  author={Bulling, Andreas and Blanke, Ulf and Schiele, Bernt},
  journal={ACM Computing Surveys (CSUR)},
  volume={46},
  number={3},
  pages={1--33},
  year={2014},
  publisher={ACM New York, NY, USA}
}

@InProceedings{10.1007/978-981-96-2468-3_16,
author="Han, Zhengxuan
and He, Mingshu",
editor="Sun, Fuchun
and Wang, Hesheng
and Long, Han
and Wei, Yifei
and Yu, Hongqi",
title="An Autoencoder Framework for Few-Shot Human Activity Recognition with Sensor Data",
booktitle="Proceedings of the 3rd International Conference on Machine Learning, Cloud Computing and Intelligent Mining (MLCCIM2024)",
year="2025",
publisher="Springer Nature Singapore",
address="Singapore",
pages="177--195",
isbn="978-981-96-2468-3"
}

@article{ganesha2024few,
  title={Few-shot transfer learning for wearable IMU-based human activity recognition},
  author={Ganesha, HS and Gupta, Rinki and Gupta, Sindhu Hak and Rajan, Sreeraman},
  journal={Neural Computing and Applications},
  volume={36},
  number={18},
  pages={10811--10823},
  year={2024},
  publisher={Springer}
}

@article{feng2019few,
  title={Few-shot learning-based human activity recognition},
  author={Feng, Siwei and Duarte, Marco F},
  journal={Expert Systems with Applications},
  volume={138},
  pages={112782},
  year={2019},
  publisher={Elsevier}
}

@article{wanyan2024comprehensive,
  title={A comprehensive review of few-shot action recognition},
  author={Wanyan, Yuyang and Yang, Xiaoshan and Dong, Weiming and Xu, Changsheng},
  journal={arXiv preprint arXiv:2407.14744},
  year={2024}
}

@inproceedings{ruan2024advances,
  title={Advances in few-shot action recognition: A comprehensive review},
  author={Ruan, Zanxi and Wei, Yingmei and Yuan, Yifei and Li, Yu and Guo, Yanming and Xie, Yuxiang},
  booktitle={2024 7th International conference on artificial intelligence and big data (ICAIBD)},
  pages={390--398},
  year={2024},
  organization={IEEE}
}

@inproceedings{malekzadeh2018protecting,
  title={Protecting sensory data against sensitive inferences},
  author={Malekzadeh, Mohammad and Clegg, Richard G and Cavallaro, Andrea and Haddadi, Hamed},
  booktitle={Proceedings of the 1st Workshop on Privacy by Design in Distributed Systems},
  pages={1--6},
  year={2018}
}

@article{khosla2020supervised,
  title={Supervised contrastive learning},
  author={Khosla, Prannay and Teterwak, Piotr and Wang, Chen and Sarna, Aaron and Tian, Yonglong and Isola, Phillip and Maschinot, Aaron and Liu, Ce and Krishnan, Dilip},
  journal={Advances in neural information processing systems},
  volume={33},
  pages={18661--18673},
  year={2020}
}

@article{diraco2023human,
  title={Human action recognition in smart living services and applications: context awareness, data availability, personalization, and privacy},
  author={Diraco, Giovanni and Rescio, Gabriele and Caroppo, Andrea and Manni, Andrea and Leone, Alessandro},
  journal={Sensors},
  volume={23},
  number={13},
  pages={6040},
  year={2023},
  publisher={MDPI}
}

@article{caputo2022you,
  title={You can't always get what you want: Towards user-controlled privacy on android},
  author={Caputo, Davide and Pagano, Francesco and Bottino, Giovanni and Verderame, Luca and Merlo, Alessio},
  journal={IEEE Transactions on Dependable and Secure Computing},
  volume={20},
  number={2},
  pages={975--987},
  year={2022},
  publisher={IEEE}
}

@inproceedings{chowdhury2025zerohar,
  title={ZeroHAR: Sensor Context Augments Zero-Shot Wearable Action Recognition},
  author={Chowdhury, Ranak Roy and Kapila, Ritvik and Panse, Ameya and Zhang, Xiyuan and Teng, Diyan and Kulkarni, Rashmi and Hong, Dezhi and Gupta, Rajesh K and Shang, Jingbo},
  booktitle={Proceedings of the AAAI Conference on Artificial Intelligence},
  volume={39},
  number={15},
  pages={16046--16054},
  year={2025}
}

@article{yang2024privacy,
  title={Privacy-preserving human activity sensing: A survey},
  author={Yang, Yanni and Hu, Pengfei and Shen, Jiaxing and Cheng, Haiming and An, Zhenlin and Liu, Xiulong},
  journal={High-Confidence Computing},
  volume={4},
  number={1},
  pages={100204},
  year={2024},
  publisher={Elsevier}
}

@article{kumar2022contrastive,
  title={Contrastive self-supervised learning: review, progress, challenges and future research directions},
  author={Kumar, Pranjal and Rawat, Piyush and Chauhan, Siddhartha},
  journal={International Journal of Multimedia Information Retrieval},
  volume={11},
  number={4},
  pages={461--488},
  year={2022},
  publisher={Springer}
}

@article{nguyen2021contrastive,
  title={Contrastive learning for neural topic model},
  author={Nguyen, Thong and Luu, Anh Tuan},
  journal={Advances in neural information processing systems},
  volume={34},
  pages={11974--11986},
  year={2021}
}

@article{zhang2022tn,
  title={TN-ZSTAD: Transferable network for zero-shot temporal activity detection},
  author={Zhang, Lingling and Chang, Xiaojun and Liu, Jun and Luo, Minnan and Li, Zhihui and Yao, Lina and Hauptmann, Alex},
  journal={IEEE Transactions on Pattern Analysis and Machine Intelligence},
  volume={45},
  number={3},
  pages={3848--3861},
  year={2022},
  publisher={IEEE}
}

@inproceedings{zhang2020zstad,
  title={Zstad: Zero-shot temporal activity detection},
  author={Zhang, Lingling and Chang, Xiaojun and Liu, Jun and Luo, Minnan and Wang, Sen and Ge, Zongyuan and Hauptmann, Alexander},
  booktitle={Proceedings of the IEEE/CVF Conference on Computer Vision and Pattern Recognition},
  pages={879--888},
  year={2020}
}

@inproceedings{nag2023semantics,
  title={Semantics guided contrastive learning of transformers for zero-shot temporal activity detection},
  author={Nag, Sayak and Goldstein, Orpaz and Roy-Chowdhury, Amit K},
  booktitle={Proceedings of the IEEE/CVF Winter Conference on Applications of Computer Vision},
  pages={6243--6253},
  year={2023}
}

@inproceedings{cheng2013towards,
  title={Towards zero-shot learning for human activity recognition using semantic attribute sequence model},
  author={Cheng, Heng-Tze and Griss, Martin and Davis, Paul and Li, Jianguo and You, Di},
  booktitle={Proceedings of the 2013 ACM international joint conference on Pervasive and ubiquitous computing},
  pages={355--358},
  year={2013}
}

@article{tong2021zero,
  title={Zero-shot learning for imu-based activity recognition using video embeddings},
  author={Tong, Catherine and Ge, Jinchen and Lane, Nicholas D},
  journal={Proceedings of the ACM on Interactive, Mobile, Wearable and Ubiquitous Technologies},
  volume={5},
  number={4},
  pages={1--23},
  year={2021},
  publisher={ACM New York, NY, USA}
}

@article{al2020zero,
  title={Zero-shot human activity recognition using non-visual sensors},
  author={Al Machot, Fadi and R. Elkobaisi, Mohammed and Kyamakya, Kyandoghere},
  journal={Sensors},
  volume={20},
  number={3},
  pages={825},
  year={2020},
  publisher={MDPI}
}

@article{xia2024ts2act,
  title={TS2ACT: Few-shot human activity sensing with cross-modal co-learning},
  author={Xia, Kang and Li, Wenzhong and Gan, Shiwei and Lu, Sanglu},
  journal={Proceedings of the ACM on Interactive, Mobile, Wearable and Ubiquitous Technologies},
  volume={7},
  number={4},
  pages={1--22},
  year={2024},
  publisher={ACM New York, NY, USA}
}

@article{civitarese2024large,
  title={Large language models are zero-shot recognizers for activities of daily living},
  author={Civitarese, Gabriele and Fiori, Michele and Choudhary, Priyankar and Bettini, Claudio},
  journal={arXiv preprint arXiv:2407.01238},
  year={2024}
}

@article{chhetri2022user,
  title={User-centric privacy controls for smart homes},
  author={Chhetri, Chola and Genaro Motti, Vivian},
  journal={Proceedings of the ACM on Human-Computer Interaction},
  volume={6},
  number={CSCW2},
  pages={1--36},
  year={2022},
  publisher={ACM New York, NY, USA}
}

@article{asif2018increasing,
  title={Increasing user controllability on device specific privacy in the Internet of Things},
  author={Asif, Waqar and Rajarajan, Muttukrishnan and Lestas, Marios},
  journal={Computer Communications},
  volume={116},
  pages={200--211},
  year={2018},
  publisher={Elsevier}
}

@inproceedings{alharbi2020synthetic,
  title={Synthetic sensor data for human activity recognition},
  author={Alharbi, Fayez and Ouarbya, Lahcen and Ward, Jamie A},
  booktitle={2020 International Joint Conference on Neural Networks (IJCNN)},
  pages={1--9},
  year={2020},
  organization={IEEE}
}

@inproceedings{norgaard2018synthetic,
  title={Synthetic sensor data generation for health applications: A supervised deep learning approach},
  author={Norgaard, Skyler and Saeedi, Ramyar and Sasani, Keyvan and Gebremedhin, Assefaw H},
  booktitle={2018 40th Annual International Conference of the IEEE Engineering in Medicine and Biology Society (EMBC)},
  pages={1164--1167},
  year={2018},
  organization={IEEE}
}

@article{thukral2025cross,
  title={Cross-Domain HAR: Few-Shot Transfer Learning for Human Activity Recognition},
  author={Thukral, Megha and Haresamudram, Harish and Ploetz, Thomas},
  journal={ACM Transactions on Intelligent Systems and Technology},
  volume={16},
  number={1},
  pages={1--35},
  year={2025},
  publisher={ACM New York, NY}
}

@inproceedings{deng2020few,
  title={Few-shot human activity recognition on noisy wearable sensor data},
  author={Deng, Shizhuo and Hua, Wen and Wang, Botao and Wang, Guoren and Zhou, Xiaofang},
  booktitle={Database Systems for Advanced Applications: 25th International Conference, DASFAA 2020, Jeju, South Korea, September 24--27, 2020, Proceedings, Part II 25},
  pages={54--72},
  year={2020},
  organization={Springer}
}

@inproceedings{chathoth2022differentially,
  title={Differentially private federated continual learning with heterogeneous cohort privacy},
  author={Chathoth, Ajesh Koyatan and Necciai, Clark P and Jagannatha, Abhyuday and Lee, Stephen},
  booktitle={2022 IEEE International Conference on Big Data (Big Data)},
  pages={5682--5691},
  year={2022},
  organization={IEEE}
}

@article{chathoth2021federated,
  title={Federated intrusion detection for iot with heterogeneous cohort privacy},
  author={Chathoth, Ajesh Koyatan and Jagannatha, Abhyuday and Lee, Stephen},
  journal={arXiv preprint arXiv:2101.09878},
  year={2021}
}

@inproceedings{chathoth2024dynamic,
  title={Dynamic black-box backdoor attacks on iot sensory data},
  author={Chathoth, Ajesh Koyatan and Lee, Stephen},
  booktitle={2024 IEEE 6th International Conference on Trust, Privacy and Security in Intelligent Systems, and Applications (TPS-ISA)},
  pages={182--191},
  year={2024},
  organization={IEEE}
}

@article{chathoth2025pcap,
  title={PCAP-Backdoor: Backdoor Poisoning Generator for Network Traffic in CPS/IoT Environments},
  author={Chathoth, Ajesh Koyatan and Lee, Stephen},
  journal={arXiv preprint arXiv:2501.15563},
  year={2025}
}

@inproceedings{melnyk2025hardware,
  title={Hardware Anomaly Detection in Microcontrollers Through Watchdog-Assisted Property Enforcement},
  author={Melnyk, Maksym and Thomas, Jacob and Wandera, Max and Chathoth, Ajesh Koyatan and Zuzak, Michael},
  booktitle={2025 IEEE International Conference on Consumer Electronics (ICCE)},
  pages={1--6},
  year={2025},
  organization={IEEE}
}

@article{hassija2019survey,
  title={A survey on IoT security: application areas, security threats, and solution architectures},
  author={Hassija, Vikas and Chamola, Vinay and Saxena, Vikas and Jain, Divyansh and Goyal, Pranav and Sikdar, Biplab},
  journal={IEEe Access},
  volume={7},
  pages={82721--82743},
  year={2019},
  publisher={IEEE}
}

\end{document}